\documentclass[review,11pt]{ReportTemplate}
\usepackage{setspace} 
\usepackage{graphicx} 
\usepackage{subfigure} 
\usepackage{times}
\usepackage[utf8]{inputenc} 
\usepackage[T1]{fontenc}    
\usepackage{nicefrac}       
\usepackage[numbers]{natbib}

\usepackage{caption}
\usepackage[all]{xy}
\usepackage{wrapfig}
\usepackage{amsmath,amssymb}

\usepackage{amsfonts}
\usepackage{paralist}
\usepackage{multirow}

\usepackage{graphics}
\usepackage{color}
\usepackage{url}
\usepackage{array}
\usepackage{verbatim}
\usepackage{microtype}
\usepackage[american]{babel}
\usepackage{makecell}
\usepackage{hyperref}
\usepackage[normalem]{ulem}

\usepackage{amsmath,amsfonts,bm}

\def\eqref#1{equation~\ref{#1}}

\def\1{\bm{1}}

\def\vf{{\bm{f}}}
\def\vg{{\bm{g}}}
\def\vh{{\bm{h}}}

\def\vw{{\bm{w}}}

\DeclareMathAlphabet{\mathsfit}{\encodingdefault}{\sfdefault}{m}{sl}
\SetMathAlphabet{\mathsfit}{bold}{\encodingdefault}{\sfdefault}{bx}{n}

\newcommand{\Var}{\mathrm{Var}}

\DeclareMathOperator*{\argmax}{arg\,max}
\DeclareMathOperator*{\argmin}{arg\,min}

\usepackage{array} 
\usepackage{bigstrut,multirow,rotating}
\usepackage{booktabs} 
\usepackage{longtable}
\usepackage{bm}

\usepackage{algorithm}
\usepackage{algorithmic}

\theoremstyle{definition}
\newtheorem{theorem}{Theorem}
\newtheorem{lemma}{Lemma}

\newcommand{\bproof}{\bigskip {\bf Proof. }}

\begin{document}

\begin{frontmatter}	
	\title{Forest Representation Learning Guided by Margin Distribution}
	
	\author{Shen-Huan Lv}
	\author{Liang Yang}
	\author{Zhi-Hua Zhou\corref{cor1}}
	\address{National Key Laboratory for Novel Software Technology\\
		Nanjing University, Nanjing 210023, China} \cortext[cor1]{\small Corresponding author.
		Email: zhouzh@nju.edu.cn}
	
	\begin{abstract}
		In this paper, we reformulate the forest representation learning approach as an additive model which boosts the augmented feature instead of the prediction. We substantially improve the upper bound of generalization gap from $\mathcal{O}(\sqrt\frac{\ln m}{m})$ to $\mathcal{O}(\frac{\ln m}{m})$, while $\lambda$ - the margin ratio between the margin standard deviation and the margin mean is small enough. This tighter upper bound inspires us to optimize the margin distribution ratio $\lambda$. Therefore, we design the margin distribution reweighting approach (mdDF) to achieve small ratio $\lambda$ by boosting the augmented feature. Experiments and visualizations confirm the effectiveness of the approach in terms of performance and representation learning ability. This study offers a novel understanding of the cascaded deep forest from the margin-theory perspective and further uses the mdDF approach to guide the layer-by-layer forest representation learning.
	\end{abstract}
	
	
\end{frontmatter}

\section{Introduction}

In recent years, deep neural networks have achieved excellent performance in many application scenarios such as face recognition and automatic speech recognition (ASR) \citep{lecun15deep}. However, deep neural networks are difficult to be interpreted. This defect severely restricts the development of deep learning in a few application scenarios, where the model's interpretability is needed. Moreover, the deep neural networks are very data-hungry due to the large complexity of the models, which means that the model's performance can decrease significantly when the size of the training data decreases \citep{elsayed18large,lv18optimal}.  

In many real tasks, due to the high cost of data collection and labeling, the amount of labeled training data may be insufficient to train a deep neural network. In such a situation, traditional learning methods such as random forest (R.F.) \citep{breiman01random}, gradient boosting decision tree (GBDT) \citep{friedman01greedy,chen16xgboost}, support-vector machines (SVMs) \citep{cortes95svm}, etc., are still good choices. By realizing that the essence of deep learning lies in the layer-by-layer processing, in-model feature transformation, and sufficient model complexity \citep{zhou18deep}, recently \citet{zhou17deep} proposed the deep forest model and the gcForest algorithm to achieve \emph{forest representation learning}. It can achieve excellent performance on a broad range of tasks, and can even perform well on small or middle-scale of data. Later on, a more efficient improvement was presented \citep{pang18improving}, and it shows that forest is able to do auto-encoder which thought to be a specialty of neural networks \citep{feng18encoder}. The tree-based multi-layer model can even do distributed representation learning which was thought to be a special feature of neural networks \citep{feng18multi}. \citet{utkin18siamese} proposed a Siamese deep forest as an alternative to the Siamese neural networks to solve the metric learning tasks.

Though deep forest has achieved great success, its theoretical exploration is less developed. The layer-by-layer representation learning is important for the cascaded deep forest, however, the cascade structure in deep forest models does not have a sound interpretation.  We attempt to explain the benefits of the cascaded deep forest in the view of boosted representations.

\subsection{Our results}
In Section~\ref{formulate}, we reformulate the cascade deep forest as an additive model (strong classifier) optimizing the margin distribution:
\begin{equation}\label{eq:f}
F(x) =  \sum_{t=1}^{T} \alpha_t h_t\left([x,f_{t-1}(x)]\right) ,
\end{equation}
where $\alpha_t$ is a scalar determined by $\ell_{\text{md}}$ - the margin distribution loss function reweighting the training samples. The input of forest block $h_t$ are the \textit{raw feature} $x$ and the \textit{augmented feature} $f_{t-1}=\sum_{l=1}^{t-1}\alpha_l h_l$: 
\begin{equation}
h_t(x)=g_t\left([x,f_{t-1}(x)]\right)=g_t\left(\left[x,\sum_{l=1}^{t-1}\alpha_l h_l(x)\right]\right),
\end{equation}
which is defined by such a recursion form. Unlike all the weak classifiers of \textit{traditional boosting} are chosen from the same hypotheses set $\mathcal{H}$, the layer-$t$ hypotheses set in the cascade deep forest contains that of the previous layer, i.e., $\mathcal{H}_{t-1} \subset \mathcal{H}_{t}, \forall t\geq 2$, due to $h_t$ is recursive. We name such a cascaded representation learning algorithm \textit{margin distribution deep forest} (mdDF).

In Section~\ref{theory}, we give a new upper bound on the generalization error of such an additive model:
\begin{equation}
	\Pr_{D}[y F(x)<0]-\Pr_S[yF(x)<r]\leq\frac{\ln \sum_{t=1}^{T} \alpha_{t}\left|\mathcal{H}_{t}\right|}{r^2}\cdot\frac{\ln m}{m} + \lambda\sqrt{\frac{\ln \sum_{t=1}^{T} \alpha_{t}\left|\mathcal{H}_{t}\right|}{r^2}\cdot\frac{\ln m}{m}},
\end{equation}
where $m$ is the size of training set, $r$ is a margin parameter, $\lambda=\sqrt{\frac{\Var [yF(x)]}{\mathbb{E}_S^2[yF(x)]}}$ is a ratio between the margin standard deviation and the expected margin, $yF(x)$ denotes the margin of the samples.

\textbf{Margin distribution.} We prove that the generalization error can be bounded by  $\mathcal{O}(\frac{\ln m}{m}+\lambda\sqrt{\frac{\ln m}{m}})$. When the margin distribution ratio $\lambda$ is small enough, our bound will be dominated by the higher order term $\mathcal{O}(\frac{\ln m}{m})$. This bound is tighter than previous bounds proved by Rademacher's complexity $\mathcal{O}(\sqrt{\frac{\ln m}{m}})$ \citep{cortes14deep}. This result inspires us to optimize the margin distribution by minimizing the ratio $\lambda$. Therefore, we utilize an appropriate margin distribution loss function $\ell_{\text{md}}$ to optimize the first- and second-order statistics of margin.

\textbf{Mixture coefficients.} As for the overfitting risk of such a deep model, our bound inherits the conclusion in \citet{cortes14deep}. The cardinality of hypotheses set $\mathcal{H}$ is controlled by the mixture coefficients $\alpha_t$s in \eqref{eq:f}. The hypotheses-set term $\sum_{t=1}^T\alpha_{t}|\mathcal{H}_{t}|$ in our bound implies that, while some hypothesis sets used for learning could have a large complexity, this may not be detrimental to generalization if the corresponding total mixture weight is relatively small. In other words, the coefficients $\alpha_{t}$s need to minimize the expected margin distribution loss $\mathbb{E}_{x\sim S}[\ell_{md}\left(\sum_{l=1}^{t}\alpha_l \gamma_{l}(x)\right)]$, which implies the generalization ability of the $t$ layer cascaded deep forest.

Extensive experiments validate that mdDF can effectively improve the performance on classification tasks, especially for categorical and mixed modeling tasks. More intuitively, the visualizations of the learned features in Figure~\ref{fig:tsne} and Figure~\ref{fig:curve} show great in-model feature transformation of the mdDF algorithm. The mdDF not only inherits all merits from the \textit{cascaded deep forest} but also boosts the learned features over layers in cascade forest structure.

\subsection{Additional related work}
The gcForest \citep{zhou18deep} is constructed by multi-grained scanning operation and cascade forest structure. The multi-grained scanning operation aims to dealing with the raw data which holds spatial or sequential relationships. The cascade forest structure aims to achieving in-model feature transformation, i.e., layer-by-layer representation learning. It can be viewed as an ensemble approach that utilizes almost all categories of strategies for diversity enhancement, e.g., input feature manipulation and output representation manipulation \citep{zhou12ensemble}. 

\citet{krogh95neural} have given a theoretical equation derived from error-ambiguity decomposition:
\begin{equation} \label{diversity}
E=\bar{E}-\bar{A},
\end{equation}
where $E$ denotes the error of an ensemble, $\bar{E}$ denotes the average error of individual classifiers in the ensemble, and $\bar{A}$ denotes the average ambiguity, later called diversity, among the individual classifiers. This offers general guidance for ensemble construction, however, it cannot be taken as an objective function for optimization, because the ambiguity is mathematically defined in the derivation and cannot be operated directly.

In this paper, we will use the margin distribution theory to analyze the cascade structure in deep forest and guide its layer-by-layer representation learning. The margin theory was first used to explain the generalization of the Adaboost algorithm \citep{schapire98boosting,breiman99arc}. Then a sequence of research \citep{reyzin06boosting,wang11boosting,gao13boosting} tries to prove the relationship between the generalization gap and the empirical margin distribution for boosting algorithm. \citet{cortes17adanet} propose a deep boosting algorithm which boosts the accuracy of variant depth decision trees, and \citet{cortes17adanet,huang18learning} offer a Rademacher bcomplexity analysis to deep neural networks. However, these theoretical results depend on the Rademacher complexity rather than the margin distribution. Since the Rademacher complexity of the forest module cannot be explicitly formulized, it cannot be taken as an objective function for optimization.

\section{Cascaded Deep Forest} \label{formulate}

\begin{figure}[t]
	\begin{center}
		\centerline{\includegraphics[width=\textwidth]{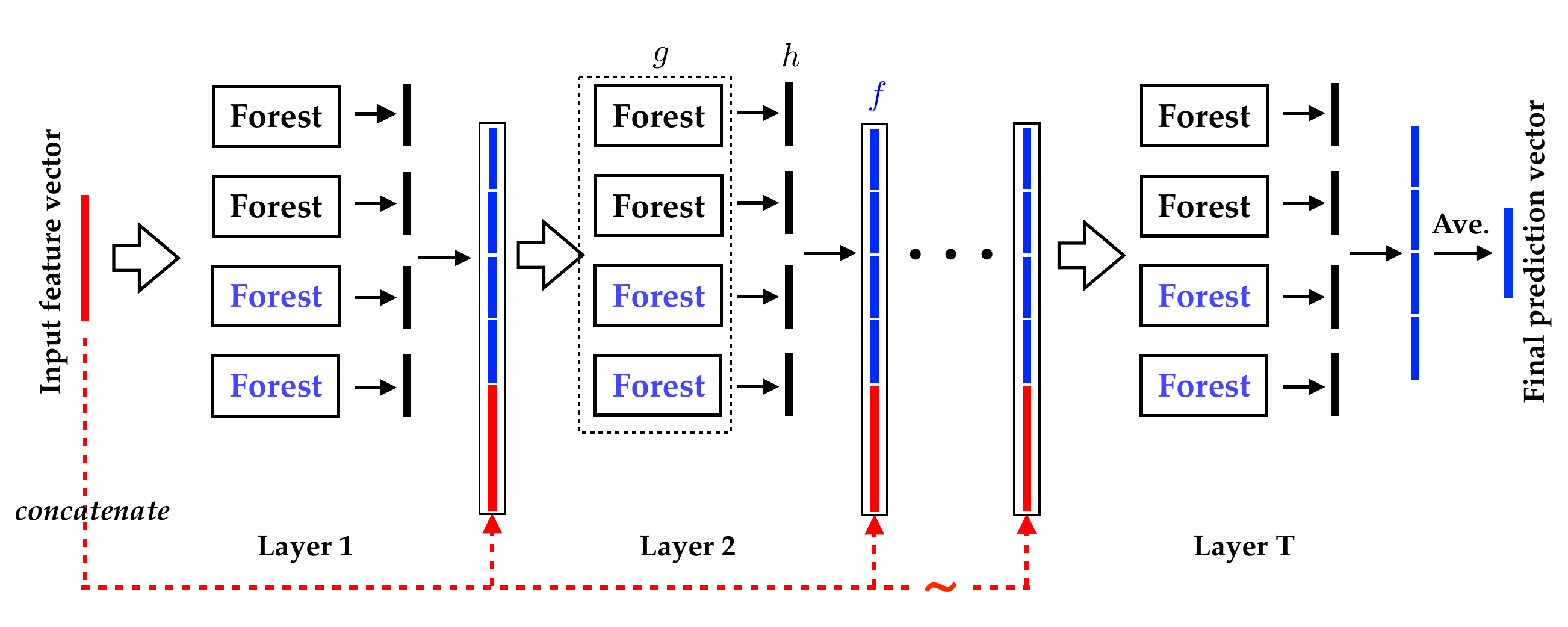}}
		\caption{Standard cascade structure of deep forest model can be viewed as a layer-by-layer process through concatenating the prediction vector with the input feature vector, which is called "preconc". This feature augmentation can achieve feature enrichment.}
		\label{structure}
	\end{center}
	\vskip -0.0in
\end{figure}

As shown in Figure~\ref{structure}, the \textit{cascaded deep forest} is composed of stacked entities referred to as forest block $g_t$s. Each forest block consists of several forest modules, which are commonly RF (random forest) \citep{breiman01random} and CRF (Completely-random forest) \citep{zhou17deep}. Cascade structure transmits the samples' representation layer-by-layer by concatenating the \textit{augmented feature} $f$ onto the origin input feature $x$. In fact, we can name this operation ``preconc" (prediction concatenation), because the \textit{augmented feature} is the prediction scores of forests in each layer. It is worth noting that ``preconc" is completely different from the stacking operation \citep{wolpert92stacked,breiman96stacked} in traditional ensemble learning. The second-level learners in stacking act on the prediction space composed of different base learners and the information of origin input feature space is ignored. Using the stacking operation with more than two layers would suffer seriously from overfitting in the experiment, and cannot enable a deep model by itself. The cascade structure is the key to success of forest representation learning, however, there has been no explicit explanation for this layer-by-layer process yet. 

Firstly we reformulate the \textit{cascaded deep forest} as an additive model mathematically in this section. We consider training and test samples generated i.i.d. from distribution $\mathcal{D}$ over $\mathcal{X}\times\mathcal{Y}$, where $\mathcal{X} \in \mathbb{R}^n$ is the input space and $\mathcal{Y} \in \{1,2,\dots,s\}$ is the label space. We denote by $S$ a training set of $m$ samples drawn according to $\mathcal{D}^m$. $\mathcal{H}_1,\mathcal{H}_2,\dots,\mathcal{H}_T$ denote $T$ families ordered by increasing complexity, i.e., $\mathcal{H}_1 \subset \mathcal{H}_2 \dots \subset \mathcal{H}_T $.

A \textit{cascaded deep forest} algorithm can be formalized as follows. We use a quadruple form $(\vg, \vf, \bm{\mathcal{D}},\vh)$ where

	\qquad $\bullet$ \quad$\vg=(g_1,g_2,\dots,g_T)$, where $g_t$ denotes the function computed by the $t$-th forest block which is defined by \eqref{g};
	
	\qquad $\bullet$ \quad$\vh=(h_1,h_2,\dots,h_T)$, where $h_t$ denotes the $t$ layer cascaded forest defined by \eqref{h}, and $h_t$ drawn from the hypothesis set $\mathcal{H}_t$;
	
	\qquad $\bullet$ \quad$\vf=(f_1,f_2,\dots,f_T)$, where $f_t$ denotes the \textit{augmented feature} in layer $t$, which is defined by \eqref{eq1};
	
	\qquad $\bullet$ \quad $\bm{\mathcal{D}}=(\mathcal{D}_1,\mathcal{D}_2,\dots,\mathcal{D}_T)$, where $\mathcal{D}_t$ is the updated sample distribution in layer $t$.

$g_t$ is the $t$-level weak module returned by the random forest block algorithm \ref{rfb}. It is learned from the \textit{raw training sample} $S=\{(x_1,y_1),\dots,(x_m,y_m)\}$ and the \textit{augmented feature} from the previous layer $f_{t-1}(x_i), i\in[m]$ and the reweighting distribution $\mathcal{D}_t$: 
\begin{equation} \label{g}
g_t=
\begin{cases}
\mathcal{A}_{\text{rfb}}\left([x_i;y_i]_{i=1}^m,\mathcal{D}\right) & t=1,\\
\mathcal{A}_{\text{rfb}}\left([x_i,f_{t-1}(x_i);y_i]_{i=1}^m,\mathcal{D}_t\right) & t>1.
\end{cases}
\end{equation}
With these weak modules, we can define the $t$ layer \textit{cascaded deep forest} as:
\begin{equation}\label{h}
h_t(x)=
\begin{cases}
g_t(x) & t = 1,\\
g_t\left([x,f_{t-1}(x)]\right) & t > 1,
\end{cases}
\end{equation}

\begin{algorithm}[t] 
	\caption{Random forest block $\mathcal{A}_{\text{rfb}}$ [\citet{zhou18deep}]}
	\label{rfb}
	\renewcommand{\algorithmicrequire}{\bfseries Input:}
	\renewcommand{\algorithmicensure}{\bfseries Output:}
	\begin{algorithmic}[1]
		\REQUIRE {Re-sample a training set $S$ with the distribution $\mathcal{D}_t$ and augmented feature $f_{t-1}(x_i), \forall i\in [m]$.}
		\ENSURE {The $t$-level forest block module $g_t$.}
		\STATE {Divide $S$ to k-fold subsets $\{S_1,\dots,S_k\}$ randomly.}
		\FOR {$S_i$ in $\{S_1,S_2,\dots,S_k\}$}
		\STATE {Using $S/S_i$ to train two random forests and two completely random forests.}
		\STATE {Compute the prediction rate $p_t^i(j)$ for the $j$-th leaf node generated by $S/S_i$.}
		\STATE {$g_t([x,f_{t-1}(x)])\leftarrow \mathbb{E}_{j}[p_t^i(j)]$, for any training sample $(x,y)\in S_i$.}
		\ENDFOR
		\STATE {$g_t([x,f_{t-1}(x)])\leftarrow \mathbb{E}_{i,j}[p_t^i(j)]$, for any test sample $(x,y)\in \mathcal{D}$.}
		\STATE {\textbf{return} The $t$-level forest block module $g_t$.}
	\end{algorithmic}
\end{algorithm}

The \textit{augmented feature} $f_t:\mathcal{X}\rightarrow\mathcal{C}$ is defined as follows:
\begin{equation}\label{eq1}
f_t(x)=
\begin{cases}
\alpha_{t}h_t(x) & t = 1,\\
\alpha_{t}h_t\left([x,f_{t-1}(x)]\right) + f_{t-1}(x) & t > 1,
\end{cases}
\end{equation}
where $\alpha_{t}$ and $\mathcal{D}_t$ is need to be optimized and updated.

Here, we find that the $t$ layer \textit{cascaded deep forest} is defined as a recursion form:
\begin{equation}
h_t(x)=g_t\left([x,f_{t-1}(x)]\right)=g_t\left(\left[x,\sum_{l=1}^{t-1}\alpha_l h_l(x)\right]\right).
\end{equation}
Unlike all the weak classifiers of \textit{traditional boosting} are chosen from the same hypotheses set $\mathcal{H}$, the layer-$t$ hypotheses set in the \textit{cascade deep forest} contains that of the previous layer, similar to the hypotheses sets of the deep neural networks (DNNs) in different depth, i.e., $\mathcal{H}_{t-1} \subset \mathcal{H}_{t}, \forall t\geq 2$.

The entire cascaded model $\tilde{F}:\mathcal{X}\rightarrow\mathcal{Y}$ is defined as follows:
\begin{equation}\label{eq4}
\tilde{F}(x)=\tilde{\sigma}(F(x))=\argmax_{j\in\{1,2,\dots,s\}}{\left[\sum_{t=1}^{T}\alpha_t h^j_t(x)\right]},
\end{equation}
where $F(x)$ is the final prediction vector of cascaded deep forest for classification and $\tilde{\sigma}$ denotes a map from average prediction score vector to a label.

\textbf{Note.} Here we generalize the formula of \textit{cascaded deep forest} as an additive model. In fact, the original version \citep{zhou17deep} sets the data distribution invariable $\mathcal{D}_t\equiv\mathcal{D}$ and the augmented feature non-additive $f_t=h_t([x,f_{t-1}(x)])$, even the final prediction vector is the direct output $F(x)=h_T(x)$. Through the generalization analysis in next section, we will explain why we need to optimize $\alpha_t$ and update $\mathcal{D}_t$.

\section{Generalization Analysis} \label{theory}

In this section, we analyze the generalization error to understand the complexity of the \textit{cascaded deep forest} model. For simplicity, we consider the \textbf{binary classification} task. We define the strong classifier as $F(x)= \sum_{t=1}^{T} \alpha_t h_t $, i.e., the cascaded deep forest reformulated as an additive model. Now we define the margin for sample $(x,y)$ as $yF(x) \in [-1,1]$, which implies the confidence of prediction. We assume that the hypotheses set $\mathcal{H}$ of base classifiers $\{h_1,h_2,\dots,h_T\}$ can be decomposed as the union of $T$ families $\mathcal{H}_1,\mathcal{H}_2,\dots,\mathcal{H}_T$ ordered by increasing complexity, where $\forall t \geq 1, \mathcal{H}_{t} \subset \mathcal{H}_{t+1} \ \text{and}\  h_t\in \mathcal{H}_t$. Remarkably, the complexity term of these bounds admits an explicit dependency in terms of the mixture coefficients defining the ensembles. Thus, the ensemble family we consider is $\mathcal{F}=\operatorname{conv}\left(\bigcup_{t=1}^{T} \mathcal{H}_{t}\right)$, that is the family of functions $F(x)$ of the form $F(x)=\sum_{t=1}^{T} \alpha_{t} h_{t}(x)$, where $\boldsymbol{\alpha}=\left(\alpha_{1}, \dots, \alpha_{T}\right)$ is in the simplex $\Delta$.

For a fixed $\mathbf{g} = (g_1,\dots,g_T)$, any $\bm{\alpha} \in \Delta$ defines a distribution over $\{ g_1,\dots,g_T \}$. Sampling from $\{ g_1,\dots,g_T \}$ according to $\bm{\alpha}$ and averaging leads to functions $G = \frac{1}{n}\sum_{i=1}^{T}n_tg_t$ for some $\mathbf{n} = (n_1,\dots,n_T)$, with $\sum_{t=1}^{T}n_t = n$, and $g_t \in \mathcal{H}_{t}$.
For any $\mathbf{N} = (N_1,\dots,N_T)$ with $|\mathbf{N}|=n$, we consider the family of functions
\begin{equation}
\mathcal{G}_{\mathcal{F},\mathbf{N}}=\left\lbrace \left. \frac{1}{n}\sum_{k=1}^T\sum_{j=1}^{N_k} g_{k,j} \right| \forall(k,j)\in [T]\times[N_k],g_{k,j}\in \mathcal{H}_k \right\rbrace,
\end{equation}
and the union of all such families $\mathcal{G}_{\mathcal{F},n} = \bigcup_{|\mathbf{N}=n|}\mathcal{G}_{\mathcal{F},\mathbf{N}}$. For a fixed $\mathbf{N}$, the size of $\mathcal{G}_{\mathcal{F},\mathbf{N}}$ can be bounded as follows:
\begin{equation}
\begin{split}
	\ln\left|\mathcal{G}_{\mathcal{F},\mathbf{N}}\right| 
	&\leq \ln \left(\prod_{t=1}^{T}|\mathcal{H}_t|^{N_t}\right)\\
	& =  \sum_{t=1}^{T}\left(N_t\ln|\mathcal{H}_t|\right)\\
	& =  n\sum_{t=1}^{T}(\alpha_t\ln|\mathcal{H}_t|)\\
	& \leq n\ln\sum_{t=1}^{T}\alpha_t|\mathcal{H}_t|
\end{split}
\end{equation}

Our margin distribution theory is based on a new Bernstein-type bound as follows:

\begin{lemma} \label{lem3}
	For $F=\sum_{t=1}^{T}\alpha_tg_t\in\mathcal{F}$ and $G\in\mathcal{G}_{\mathcal{F},n}$, we have
	\begin{equation}
	\Pr_{S,\mathcal{G}_{\mathcal{F},n}}[yG(x)-yF(x)\geq \epsilon]\leq \exp\left( \frac{-n\epsilon^2}{2-2\mathbb{E}_S^2[yF(x)]+4\epsilon/3} \right).
	\end{equation}
\end{lemma}

\bproof
For $\lambda>0$, according to the Markov's inequality, we have
\begin{align}
\Pr_{S,\mathcal{G}_{\mathcal{F},n}}[yG(x)-yF(x)\geq \epsilon]&=\Pr_{S,\mathcal{G}_{\mathcal{F},n}}[(yG(x)-yF(x))n\lambda/2 \geq n\lambda\epsilon/2]\\
&\leq \exp\left( -\frac{\lambda n\epsilon}{2} \right)\mathbb{E}_{S,G_j\in \mathcal{G}_{\mathcal{F},n}} \left[ \exp\left( \frac{\lambda}{2}\sum_{j=1}^{n}(yG_j(x)-yF(x)) \right) \right]\\
&= \exp\left( -\frac{\lambda n\epsilon}{2} \right) \prod_{j=1}^{n} \mathbb{E}_{S,G_j\in \mathcal{G}_{\mathcal{F},n}} \left[ \exp\left( \frac{\lambda}{2}(yG_j(x)-yF(x)) \right) \right]
\end{align}
where the last inequality holds from the independent of $G_i$. Notice that $|yG_j(x)-yF(x)|\leq2$ (the margin is bounded: $yF(x)\in [-1,1]$), using Taylor's expansion, we get
\begin{align}
\mathbb{E}_{S,G_j\in \mathcal{G}_{\mathcal{F},n}} \left[ \exp\left( \frac{\lambda}{2}(yG_j(x)-yF(x)) \right) \right]&\leq 1+\mathbb{E}_{S,G_j\in \mathcal{G}_{\mathcal{F},n}}[(yG_j(x)-yF(x))^2]\frac{e^\lambda-1-\lambda}{4}\\
&\leq 1 + \mathbb{E}_S[1-(yF(x))^2]\frac{e^\lambda-1-\lambda}{4}\\
&\leq \exp(1-\mathbb{E}_S^2[yF(x)])\frac{e^\lambda-1-\lambda}{4}
\end{align}
where the last inequality holds from Jensen's inequality and $1+x\leq e^x$. Therefore, we have
\begin{equation}\label{s40}
\Pr_{S,\mathcal{G}_{\mathcal{F},n}}[yG(x)-yF(x)\geq \epsilon]\leq \exp\left( \frac{n(e^\lambda-1-\lambda)(1-\mathbb{E}_S[yF(x)])}{4} - \frac{\lambda n\epsilon}{2} \right)
\end{equation}
If $0<\lambda<3$, then we could use Taylor's expansion again to have
\begin{equation}
e^\lambda-\lambda-1=\sum_{i=2}^{\infty}\frac{\lambda^i}{i!}\leq\frac{\lambda^2}{2}\sum_{m=0}^{\infty}\frac{\lambda^m}{3^m} = \frac{\lambda^2}{2(1-\lambda/3)}.
\end{equation}
Now by picking $\lambda = \frac{\epsilon}{1/2 - \mathbb{E}_S^2[yF(x)]/2+\epsilon/3}$, we have
\begin{equation}\label{s42}
-\frac{\lambda\epsilon}{2} + \frac{\lambda^2(1-\mathbb{E}_S^2[yF(x)])}{8(1-\lambda/3)}\leq \frac{-\epsilon^2}{2-2\mathbb{E}_S^2[yF(x)]+4\epsilon/3}
\end{equation}
By Combining the \eqref{s40} and \ref{s42} together, we complete the proof. \qed

Since the gap between the margin of strong classifier $yF(x)$ and the  margin of classifiers in the union family $\mathcal{G}_{\mathcal{F},\mathbf{N}}$ is bounded by the margin mean, we can further obtain a margin distribution theorem as follows:

\begin{theorem} \label{generalization}
	Let $\mathcal{D}$ be a distribution over $\mathcal{X} \times \mathcal{Y}$ and $S$ be a sample of $m$ examples chosen independently at random according to $\mathcal{D}$. With probability at least $1-\delta$, for $r > 0$, the strong classifier $F(x)$ (depth-$T$ mdDF) satisfies that
	\begin{align}
	&\Pr_D[yF(x)<0]\leq \notag \inf_{r\in(0,1]}\left[\hat{R}+\frac{1}{m^d}+ \frac{3\sqrt{\mu}}{m^{3/2}}+ \frac{7\mu}{3m}+ \lambda\sqrt{\frac{3\mu}{m}}\right]
	\end{align}
	where 
	\begin{align*}
	\hat{R} &= \Pr_S[yF(x)<r], \qquad\qquad\qquad\qquad\qquad\qquad\\
	d &= \frac{2}{1-\mathbb{E}_S^2[yF(x)]+r/9}>2,\\
	\mu &= \ln m \ln(2\sum_{t=1}^T\alpha_{t}|\mathcal{H}_{t}|)/r^2+\ln\left(\frac{2}{\delta} \right),\\
	\lambda &=\sqrt{\Var [yF(x)]/\mathbb{E}_S^2[yF(x)]}.
	\end{align*}
\end{theorem}

\bproof

\begin{lemma} \label{lem1}
	[Chernoff bound \citep{chernoff52bound}] Let $X,X_1,\dots,X_m$ be $m+1$ i.i.d. random variables with $X\in[0,1]$. Then, for any $\epsilon>0$, we have
	\begin{align} 
	&\Pr[\frac{1}{m}\sum_{i=1}^{m}X_i\geq\mathbb{E}[X]+\epsilon]\leq \exp\left(-\frac{m\epsilon^2}{2}\right),\\
	&\Pr[\frac{1}{m}\sum_{i=1}^{m}X_i\leq\mathbb{E}[X]-\epsilon]\leq \exp\left(-\frac{m\epsilon^2}{2}\right).
	\end{align}
\end{lemma}

\begin{lemma} \label{lem2}[\citet{gao13boosting}] 
	For independent random variables $X_1,X_2,\dots,X_m (m\geq 5)$ with values in $[0,1]$, and for $\delta\in(0,1)$, with probability at least $1-\delta$ we have
	\begin{align} 
	&\frac{1}{m}\sum_{i=1}^{m}\mathbb{E}[X_i]-\frac{1}{m}\sum_{i=1}^{m}X_i\leq \sqrt{\frac{2\hat{V}_m\ln(2/\delta)}{m}}+\frac{7\ln(2/\delta)}{3m},\\
	&\frac{1}{m}\sum_{i=1}^{m}\mathbb{E}[X_i]-\frac{1}{m}\sum_{i=1}^{m}X_i\geq- \sqrt{\frac{2\hat{V}_m\ln(2/\delta)}{m}}-\frac{7\ln(2/\delta)}{3m}.
	\end{align}
	where $\hat{V}_m = \sum_{i\neq j}(X_i-X_j)^2/2m(m-1)$
	
\end{lemma}

For $F=\sum_{t=1}^{T}\alpha_tg_t\in\mathcal{F}$ and $G\in\mathcal{G}_{\mathcal{F},n}$, we have $\mathbb{E}_{G\in\mathcal{G}_{\mathcal{F},n}}[G] = F$. For $\beta>0$, the Chernoff's bound in Lemma~\ref{lem1} gives
\begin{align}
\Pr_D{[yF(x)<0]}&= \Pr_{D,\mathcal{G}_{\mathcal{F},n}}[yF(x)<0,yG(x)\geq\beta]+\Pr_{D,\mathcal{G}_{\mathcal{F},n}}[yF(x)<0,yG(x)<\beta]\\
&\leq \exp(-n\beta^2/2)+\Pr_{D,\mathcal{G}_{\mathcal{F},n}}[yG(x)<\beta]. \label{44}
\end{align}
Recall that $|\mathcal{G}_{\mathcal{F},N}|\leq \prod_{t=1}^{T}|\mathcal{H}_{t}|^{N_t}$ for a fixed $N$. Therefore, for any $\delta_n>0$, combining the union bound with Lemma~\ref{lem2} guarantees that with probability at least $1-\delta_n$ over sample $S$, for any $G\in\mathcal{G}_{\mathcal{F},N}$ and $\beta>0$
\begin{align} \label{45}
\Pr_D[yG(x)<\beta]&\leq\Pr_S[yG(x)<\beta]+\sqrt{\frac{2}{m}\hat{V}_m \ln \left(\frac{2}{\delta}\prod_{t=1}^{T}|\mathcal{H}_{t}|^{N_t}\right)} + \frac{7}{3m}\ln\left(\frac{2}{\delta}\prod_{t=1}^{T}|\mathcal{H}_{t}|^{N_t}\right)\\
&=\Pr_S[yG(x)<\beta]+\sqrt{\frac{2}{m}\hat{V}_m \sum_{i=1}^{T} N_t\ln \left(\frac{2|\mathcal{H}_{t}|}{\delta}\right)} + \frac{7}{3m}\sum_{i=1}^{T} N_t\ln \left(\frac{2|\mathcal{H}_{t}|}{\delta}\right)\\
&\leq \Pr_S[yG(x)<\beta]+\sqrt{\frac{2n}{m}\hat{V}_m  \sum_{i=1}^{T} \alpha_{t}\ln \left(\frac{2|\mathcal{H}_{t}|}{\delta}\right)} + \frac{7n}{3m}  \sum_{i=1}^{T} \alpha_{t}\ln \left(\frac{2|\mathcal{H}_{t}|}{\delta}\right) \label{29}\\
&\leq \Pr_S[yG(x)<\beta]+\sqrt{\frac{2n}{m}\hat{V}_m  \ln \left(\frac{2\sum_{i=1}^{T} \alpha_{t}|\mathcal{H}_{t}|}{\delta}\right)} + \frac{7n}{3m}  \ln \left(\frac{2\sum_{i=1}^{T} \alpha_{t}|\mathcal{H}_{t}|}{\delta}\right) \label{30}\\
\end{align}
where 
\begin{equation}
\hat{V}_m = \sum_{i\neq j}\frac{(\mathbb{I}[y_iG(x_i)<\beta]-\mathbb{I}[y_jG(x_j)<\beta])^2}{2m(m-1)},
\end{equation}

The inequality \ref{29} is a large probability bound when $n$ is large enough and inequality \ref{30} is according to the Jensen's Inequality. Since there are $T$ at most $T^n$ possible $T$-tuples $N$ with $|N|=n$, by the union bound, for any $\delta >0$, with probability at least $1-\delta$, for all $G \in \mathcal{G}_{\mathcal{F},n}$ and $\beta>0$:
\begin{equation}
\Pr_D[yG(x)<\beta]\leq \Pr_S[yG(x)<\beta]+\sqrt{\frac{2n}{m}\hat{V}_m  \ln \left(\frac{2\sum_{i=1}^{T} \alpha_{t}|\mathcal{H}_{t}|}{\delta/T^n}\right)} + \frac{7n}{3m}  \ln \left(\frac{2\sum_{i=1}^{T} \alpha_{t}|\mathcal{H}_{t}|}{\delta/T^n}\right)
\end{equation}

Meantime, we can rewrite $\hat{V}_m$
\begin{align}
\hat{V}_m &= \sum_{i\neq j}\frac{(\mathbb{I}[y_iG(x_i)<\beta]-\mathbb{I}[y_jG(x_j)<\beta])^2}{2m(m-1)}\\
&= \frac{2m^2\Pr_S[yG(x)<\beta]\Pr_S[yG(x)\geq\beta]}{2m(m-1)}\\
&= \frac{m}{m-1}\hat{V}_m^*
\end{align}
For any $\theta_1, \theta_2>0$, we utilize Chernoff's bound in Lemma~\ref{lem2} to get:
\begin{align}
\hat{V}_m^* &= \Pr_S[yG(x)<\beta]\Pr_S[yG(x)\geq\beta]\\
&\leq 3\exp(-n\theta_1^2/2) + \Pr_S[yF(x)<\beta+\theta_1]\Pr_S[yF(x)\geq\beta-\theta_1]\\
&\leq 3\exp(-n\theta_1^2/2) +\\ &\Pr_S[yF(x)<\beta+\theta_1\left|\mathbb{E}_S[yF(x)]\geq\beta+\theta_1+\theta_2\right.]\Pr_S[yF(x)\geq\beta-\theta_1|\mathbb{E}_S[yF(x)]\geq\beta+\theta_1+\theta_2]\notag\\
&\leq 3\exp(-n\theta_1^2/2) + \frac{\Var[yF(x)]}{\theta_2^2} \qquad\qquad\qquad\qquad\qquad\qquad \mbox{According to Chebyshev's Inequality}\\
&\leq 3\exp(-n\theta_1^2/2) + \frac{\Var[yF(x)]}{(\mathbb{E}_S[yF(x)]-\beta+\theta_1)^2} \\ 
&\simeq 3\exp(-n\theta_1^2/2) + \frac{\Var[yF(x)]}{\mathbb{E}_S^2[yF(x)]}       \label{52}
\end{align}
where $\Var[yF(x)] = \mathbb{E}_S[(yF(x))^2]-\mathbb{E}_S^2[yF(x)]$ is the variance of the margins.

From Lemma~\ref{lem3}, we obtain that
\begin{equation} \label{53}
\Pr_S[yG(x)<\beta]\leq\Pr_S[yF(x)<\beta+\theta_1] + \exp\left( \frac{-n\theta_1^2}{2-2\mathbb{E}_S^2[yF(x)]+4\theta_1/3} \right)
\end{equation}
Let $\theta_1=r/6$, $\beta=5r/6$ and $n = \ln m/r^2$, then we combine the \eqref{44},\ref{45},\ref{52} and \ref{53}, the proof is completed. 

\qed

\textbf{Remark 1.} From Theorem~\ref{generalization}, we know that the gap between the generalization error and empirical margin loss is generally bounded by the forests complexity $\mathcal{O}\left(\sqrt{\frac{\ln m \ln(\sum_{t=1}^T\alpha_{t}|\mathcal{H}_{t}|)}{mr^2}}\right)$, which is controlled by the ratio between the margin standard deviation and the margin mean $\lambda=\sqrt{\frac{\hat{V}_m[yF(x)]}{\mathbb{E}_S^2[yF(x)]}}$. This ratio implies that the smaller margin mean and larger margin variance can reduce the complexity of models properly, which is crucial to alleviate the overfitting problem. When the margin distribution is good enough (margin mean is large and margin variance is small), $\mathcal{O}\left(\frac{\ln m}{m}\right)$ will dominate the order of the sample complexity. This is tighter than the previous theoretical work about deep boosting \citep{cortes14deep, cortes17adanet, huang18learning} $\mathcal{O}(\sqrt{\frac{\ln m}{m}})$. 

\textbf{Remark 2.} Moreover, this novel bound inherits the property of previous bound \citep{cortes14deep}, the hypotheses term $\ln \sum_{t=1}^{T}\alpha_t|\mathcal{H}_t|$ admits an explicit dependency on the mixture coefficients $\alpha_t$s. It implies that, while some hypothesis sets used for learning could have a large complexity, this may not be detrimental to generalization if the corresponding total mixture weight is relatively small. This property also offers a potential to obtain a good generalization result through optimizing the $\alpha_t$s.

It is worth noting that the analysis here only considers the cascade structure in deep forest. Due to the simplification of the model, we do not analyze the details about the "preconc" operation and the influence of adopting a different type of forests, though these two operations play an important role in practice. The advantages of these operations are evaluated empirically in Section~\ref{experiment}.

\section{Margin Distribution Optimization}

\begin{figure}[t]
	\begin{center}
		\centerline{\includegraphics[width=0.6\textwidth]{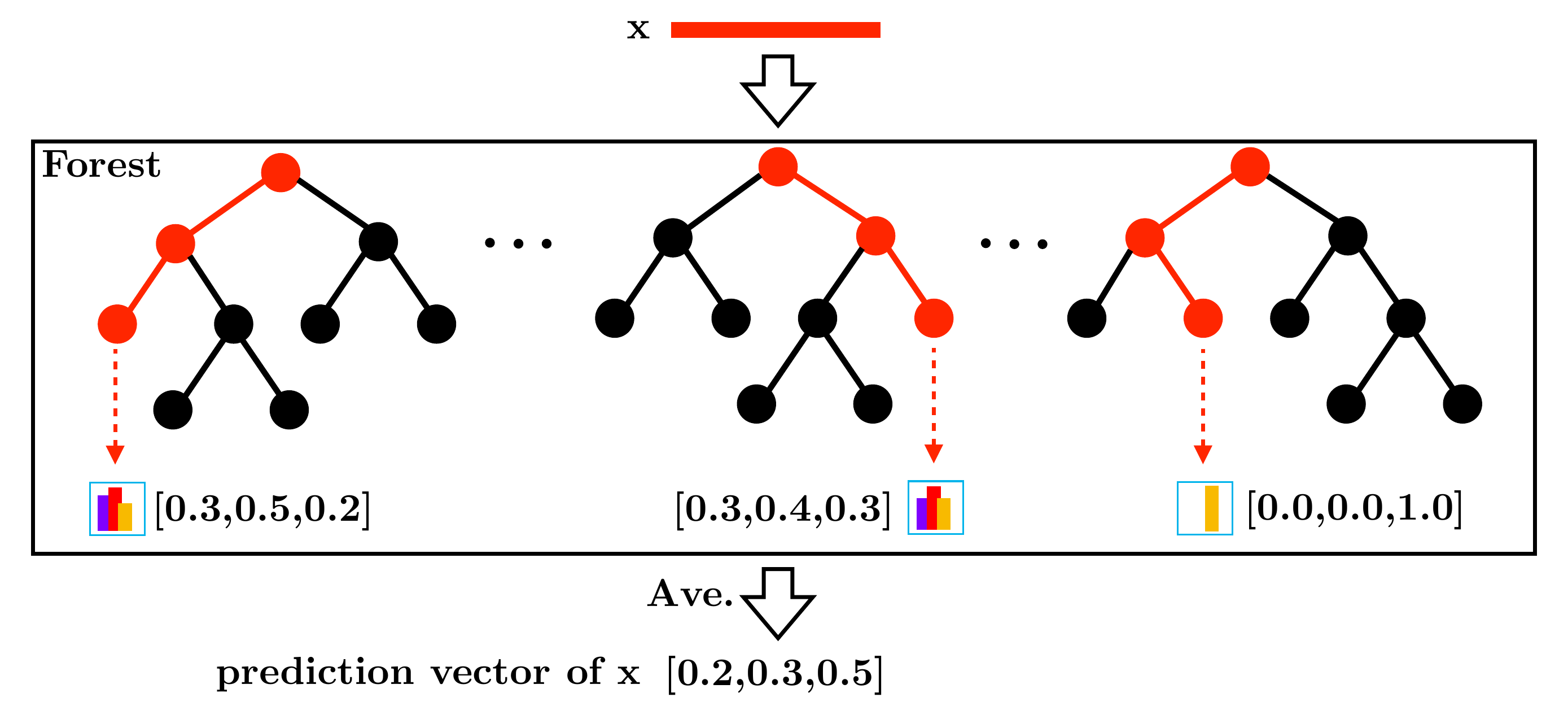}}
		\caption{ Illustration of prediction vector generation from $h_t(\cdot)$. Each component
			of the final prediction vector is an average of the outputs of individual
			trees. Different shapes in leaf nodes denote different classes.  }
		\label{fig:t2}
	\end{center}
	\vskip -0.20in
\end{figure}

The generalization theory shows the importance of optimizing the margin distribution ratio $\lambda$ and the mixture coefficients $\alpha_{t}$s. Since we reformulate the cascaded deep forest as a additive model, we utilize the reweighting approach to minimize the expected margin distribution loss 
\begin{equation}
	\mathbb{E}_{x\sim\mathcal{S}}[\ell_{md}\circ F(x)]=\mathbb{E}_{x\sim\mathcal{S}}\left[\ell_{md}\circ\sum_{t=1}^{T}\alpha_{t}h_t(x)\right],
\end{equation}
where the margin distribution loss function $\ell_{\text{md}}$ is designed to utilize the first- and second-order statistics of margin distribution. The reweighting approach helps the model to boosts the augmented feature in deeper layer, i.e., $h_t$ focus on dealing with the samples which have the large loss in the previous layers. The choice of scalar $\alpha_t$ is determined by minimizing the expected loss for $t$-layer model.

\begin{algorithm}[t] 
	\caption{mdDF}
	\label{boostlmdf}
	\renewcommand{\algorithmicrequire}{\bfseries Input:}
	\renewcommand{\algorithmicensure}{\bfseries Output:}
	\begin{algorithmic}[1] \label{al}
		\REQUIRE {Training set $S=\{(x_1,y_1),\dots,(x_m,y_m)\}$ and random forest block algorithm $\mathcal{A}_{\text{rfb}}$.}
		\ENSURE {Cascaded additive model $\tilde{F}$.}
		\STATE {Initialize $\alpha_0 \leftarrow 1, f_0\leftarrow\emptyset$}
		\STATE {Initialize sample weights: $\mathcal{D}_1(i) \leftarrow \frac{1}{m}, \forall i \in [m]$}
		\FOR {$t= 1,2,\dots,T$ } 
		\STATE {$h_t \leftarrow$ the forest block returned by $\mathcal{A}_{\text{rfb}}([x_i,f_{t-1}(x_i);y_i]_{i=1}^m,\mathcal{D}_t)$.} 
		\STATE {$\gamma_{t}(x_i) \leftarrow h_{t}^y([x_i,f_{t-1}(x_i)]) - \max_{j\neq y}h_{t}^j([x_i,f_{t-1}(x_i)]), \forall i\in[m]$.}
		\STATE {Optimize $\alpha_{t}$ using SGD algorithm.} \qquad\qquad\qquad\qquad\quad $\triangleright$\quad $\underset{\alpha_t}{\argmin} \  \mathbb{E}_{x\sim S}[\ell_{md}\left(\sum_{l=1}^{t}\alpha_l \gamma_{l}(x)\right)]$
		\STATE {$f_t(x_i)\leftarrow\alpha_{t}h_t\left([x_i,f_{t-1}(x_i)]\right) + f_{t-1}(x_i), \forall i\in [m]$.}
		\STATE {$\mathcal{D}_{t+1}(i)\leftarrow\frac{\ell_{md}\left(\sum_{l=1}^{t}\alpha_{l}\gamma_{l}(x_i)\right)}{\sum_{i=1}^{m}\ell_{md}\left(\sum_{l=1}^{t}\alpha_{l}\gamma_{l}(x_i)\right)}, \forall i\in [m]$.}
		\ENDFOR
		\STATE {\textbf{return} $\tilde{F}=\underset{j\in\{1,2,\dots,s\}}{\argmax}{\left[\sum_{t=1}^{T}\alpha_t h_t^j\right]}$.}
	\end{algorithmic}
\end{algorithm}

\subsection{Algorithm for mdDF approach.} 

We denote by $C=\mathbb{R}^s$ a prediction score space, where $s$ is the number of classes. When any sample $(x,y)\in\mathcal{D}$ passes through the cascaded deep forest model, it will get an average prediction vector in each layer: $h_{t}(x)=\left[h_{t}^1(x),h_{t}^2(x)\dots,h_{t}^s(x)\right] \in C$. According to \citet{crammer01algorithmic}, we can define the sample's margin $\gamma_{t}(x)$ for multi-class classification as:
\begin{equation}\label{eq43}
\begin{split}
	\gamma_{t}(x) &:= h_{t}^y(x) - \max_{j\neq y}h_{t}^j(x),\\
\end{split}
\end{equation}
that is, the prediction's confidence-rate. For example, in the 3-class problem, as shown in Figure~\ref{fig:t2}, the average prediction score is $[0.2, 0.3, 0.5]$, and the margin is calculated as $0.5-0.3=0.2$. 

The initial sample weights are $[1/m,1/m,\dots,1/m]$, and we update the $i$-th sample weight by:
\begin{equation}\label{eq2}
\mathcal{D}_{t+1}(i)=\frac{\ell_{md}\left(\sum_{l=1}^{t}\alpha_{l}\gamma_{l}(x_i)\right)}{\sum_{i=1}^{m}\ell_{md}\left(\sum_{l=1}^{t}\alpha_{l}\gamma_{l}(x_i)\right)},
\end{equation}
The margin distribution loss function $\ell_{md}(\cdot)$ is defined as follows:
\begin{equation}\label{mdloss}
\ell_{md}(z) = 
\begin{cases}
\frac{(z-\gamma)^2}{\gamma^2} & z \leq \gamma,\\
\frac{\mu(z-\gamma)^2}{(1-\gamma)^2} & z > \gamma,
\end{cases}
\end{equation}
where hyper-parameter $\gamma$ is a parameter as the margin mean and $\mu$ is a parameter to trade off two different kinds of deviation (keeping the balance on both sides of the margin mean). In practice, we generally choose these two hyper-parameters from the finite sets $\gamma\in\{0.7,0.75,0.8,0.85,0.9,0.95\}$ and $\mu\in\{0.01,0.05,0.1\}$. The algorithm utilizing this margin distribution optimization is summarized in Algorithm~\ref{al}. 

\subsection{The intuition of margin distribution loss function.} 

Since \citet{reyzin06boosting} found that the \textit{margin distribution} of AdaBoost is better than that of \texttt{arc-gv} \citep{breiman99arc} which is a boosting algorithm designed to maximize the minimum margin, \citet{reyzin06boosting} conjectured that margin distribution is more important to get a better generalization performance than the instance with the minimum margin. \citet{gao13boosting} prove that utilizing both the margin mean and margin variance can portray the relationship between margin and generalization performance for AdaBoost algorithm more precisely. We list the several loss functions of the algorithms based on margin theory to compare and plot them in Figure.~\ref{fig:loss}:

%
%
%

\begin{figure}[ht]
\begin{minipage}[b]{0.46\linewidth}
	\centering
	\includegraphics[width=\textwidth]{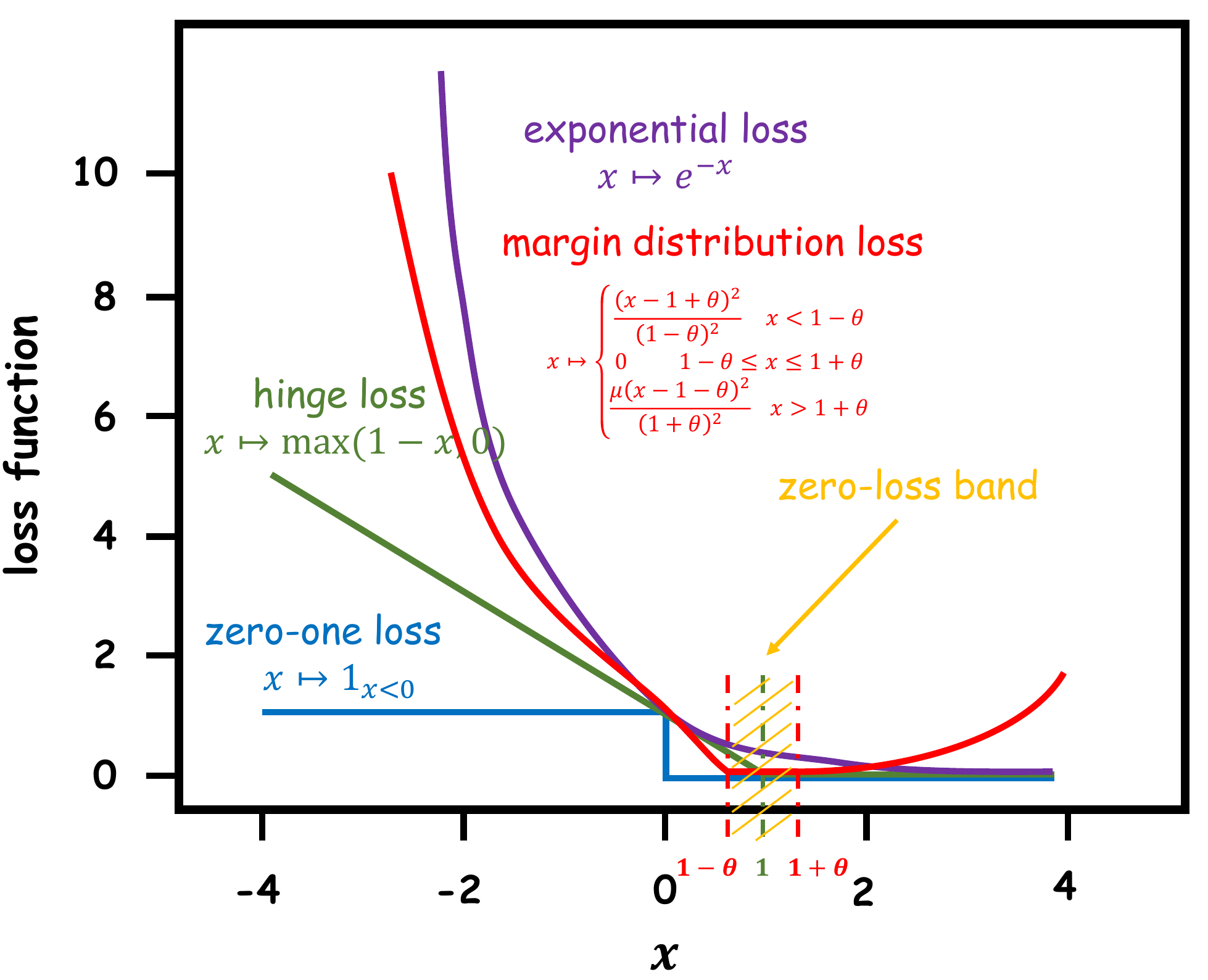}
	\caption{Several convex upper bounds of $\ell_{0-1}$.}
	\label{fig:loss}
\end{minipage}
\hfill
\begin{minipage}[b]{0.42\linewidth}
	\textbf{Exponential loss function:}
	\begin{equation}
	\ell_{\text{exp}}(x) = \exp\{-x\}.
	\end{equation} 
	
	\textbf{Hinge loss function:}
	\begin{equation}
	\ell_{\text{hinge}}(x) = \max\{1-x, 0\}.
	\end{equation} 
	
	\textbf{Margin distribution loss function:}
	\begin{equation}\label{eq:md}
	\ell_{\text{md}}(x) =
	\begin{cases}
	\frac{(x-1+\theta)^2}{(1-\theta)^2} & x \leq 1-\theta,\\
	0 & 1-\theta < x \leq 1+\theta,\\
	\frac{\mu(x-1-\theta)^2}{(1+\theta)^2} & x > 1+\theta.
	\end{cases}
	\end{equation}
\end{minipage}
\end{figure}

\begin{figure}[h]
	\centering
	\includegraphics[width=0.6\textwidth]{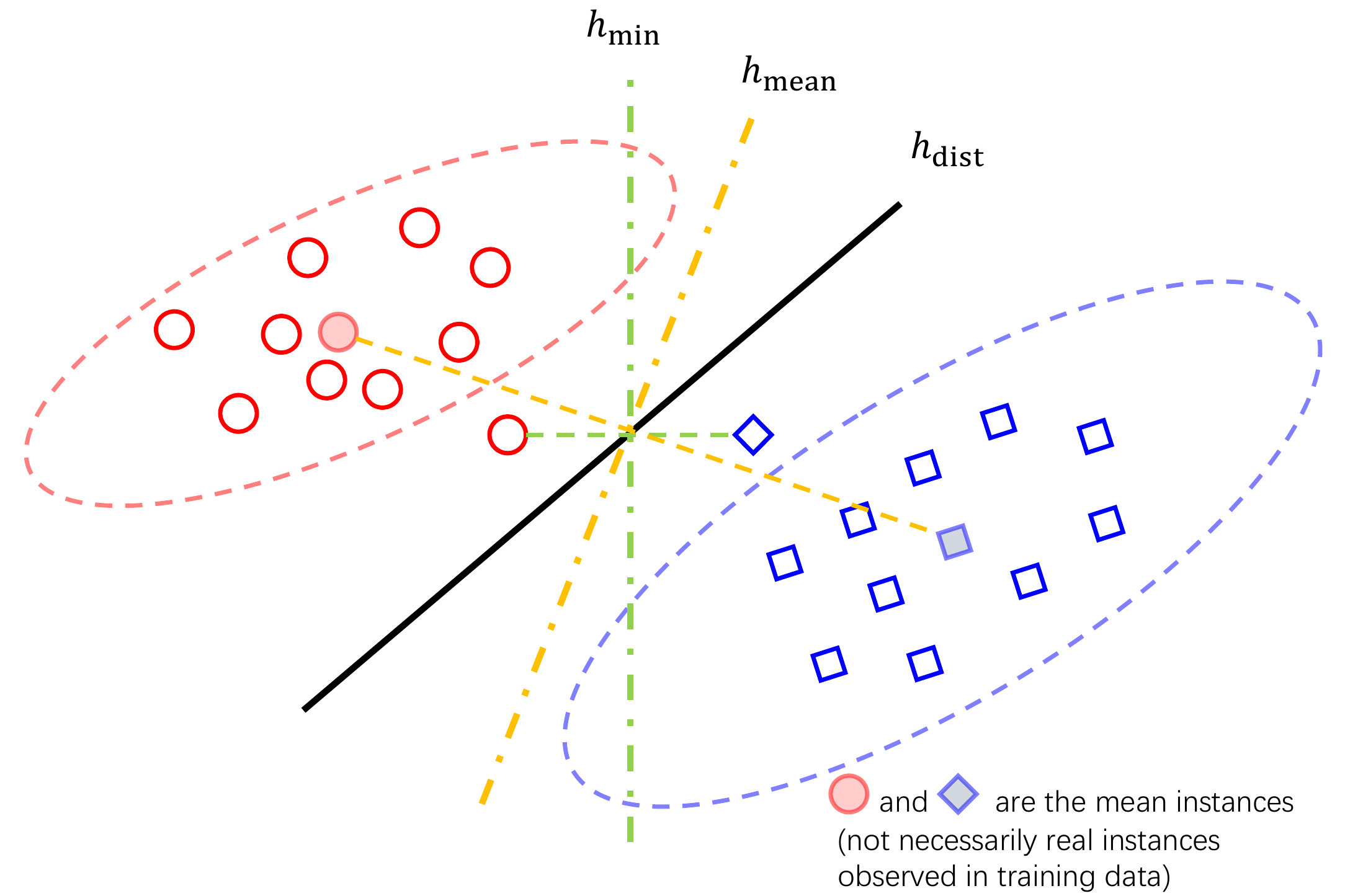}
	\caption{A simple illustration of linear separators optimizing the minimum margin, margin mean and margin distribution, respectively. $h_{\text{min}}$ represents the classifier learned by maximizing the minimum margin. $h_{\text{mean}}$ represents that learned by maximizing the margin mean. $h_{\text{dist}}$ represents that learned by optimizing the margin distribution through maximizing the margin variance and minimizing the margin variance simultaneously.}
	\label{fig:margin}
\end{figure}

\begin{figure}[h]
	\centering
	\includegraphics[width=0.6\textwidth]{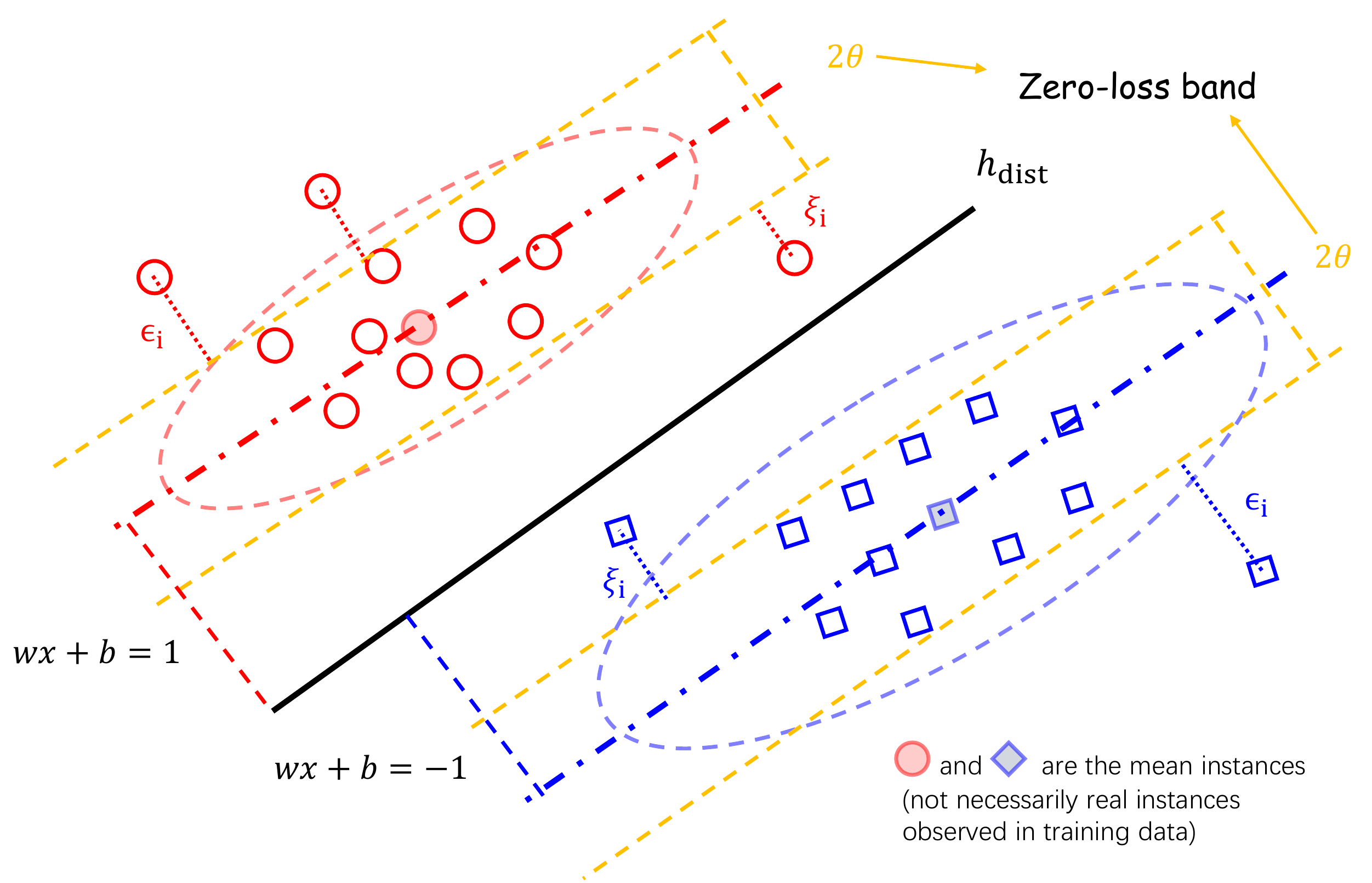}
	\caption{A simple illustration of optimal margin distribution machine (ODM). We assume that the margin mean is preset to a constant $1$, so that $\theta$ is somewhat a parameter which implies the margin variance. Since the sample points away form margin mean, i.e., $\xi_i>0 \vee \epsilon_{i}>0$, will be imposed a square-type penalty,  $[1-\theta,1+\theta]$ is the zero-loss band so as to contain as much training data as possible. At last, when we optimize the margin distribution loss with a regularization term $\Omega(\vw)$, we actually maximize the normalized margin mean $\frac{1}{\Omega(\vw)}$ with a margin variance controlled by parameter $\theta$.}
	\label{fig:odm}
\end{figure}

Compared with maximize the minimum margin (SVMs), the optimal margin distribution principle \citep{gao13boosting,zhang17odm} conjecture that maximizing the margin mean and minimizing the margin variance is the key to achieving a better generalization performance. Figure.~\ref{fig:margin} shows that optimizing the margin distribution with first- and second-order statistics can utilize more information on training data, e.g. the covariance among the different features. Inspired by this idea, \citet{zhang17odm} propose the optimal margin distribution machine (ODM) which can be formulated as \eqref{eq:odm}:

\begin{equation}\label{eq:odm}
\begin{split}
\min _{\boldsymbol{w}, \xi_{i}, \epsilon_{i}}\quad &\Omega(\boldsymbol{w})+\frac{\lambda}{m} \sum_{i=1}^{m} \frac{\xi_{i}^{2}+\mu \epsilon_{i}^{2}}{(1-\theta)^{2}}\\
\mbox{s.t.}\quad &\gamma_{h}\left(\boldsymbol{x}_{i}, y_{i}\right) \geq 1-\theta-\xi_{i}\\
&\gamma_{h}\left(\boldsymbol{x}_{i}, y_{i}\right) \leq 1+\theta+\epsilon_{i}, \forall i
\end{split}
\end{equation}

where $\theta+\xi_i$ and $\theta+\epsilon_{i}$ are the deviation of the margin $\gamma_h(x_i,y_i)$ to the margin mean, $\mu\in (0,1]$ is a parameter to trade off two different kinds of deviation (larger or less than margin mean). $\theta\in [0,1)$ is a parameter of the \textbf{zero-loss band}, which can control the number of support vectors, i.e., the sparsity of the solution, and $(1-\theta)^2$ in the denominator is to scale the second term to be a surrogate loss for 0-1 loss. Similar to support vector machines (SVMs), we can give ODM an intuitive illustration in Figure.~\ref{fig:odm}. Similar to formulating support vector machines as a combination of the hinge loss and the regularization term, we can use margin distribution loss function $\ell_{\text{md}}$ defined in \eqref{eq:md} and a regularization term to represent the ODM. Our simplified version margin distribution loss function (\ref{mdloss}) is similar to that of the ODM. Our forest representation learning approach requires as many samples as possible to train the model and generate the augmented features. Therefore, we remove the parameter $\theta$ which can control the number of support vectors. Our loss function is to optimize the margin distribution to minimize the margin distribution ratio $\lambda$, referring to Remark 2 in Section~\ref{theory}.

\section{Experiments} \label{experiment}

In this section, we provide experiments and visualizations confirmed the effectiveness of the model in terms of performance and representation learning ability. Furthermore, we conjecture that numeric modeling tasks such as image/audio data are very suitable for DNNs because their operations, such as convolution, fit well with numeric signal modeling. The deep forest model was not developed to replace DNNs for such tasks; instead, it offers an alternative when DNNs are not superior. There are plenty of tasks, especially categorical/symbolic or mixed modeling tasks, where deep forest may be found useful.

\begin{table}[h]
	\caption{Description of the dataset in terms of number of training examples, number of testing examples, number of features and number of labels.}
	\label{sample-table}
	\begin{center}
		\begin{small}
			\begin{tabular}{p{2.0cm}p{2.0cm}<{\raggedleft}p{1.6cm}<{\raggedleft}p{1.6cm}<{\raggedleft}p{1.6cm}<{\raggedleft}}
				\toprule
				\bf Dataset &\bf Attribute 					&\bf Instance	&\bf Feature  	&\bf Class 	\\
				\midrule    
				ADULT    	&Categorical					&48842			&14				&2  		\\
				YEAST    	&Categorical					&1484			&8				&10 		\\
				LETTER    	&Categorical					&20000			&16				&26 		\\
				PROTEIN		&Categorical					&24387			&357			&3			\\
				HAR         &Mixed							&10299			&561			&6  		\\
				SENSIT		&Mixed							&78823			&50				&3			\\
				SATIMAGE	&Numerical						&6435			&36				&6			\\
				MNIST		&Numerical						&70000			&784			&10			\\
				\bottomrule
			\end{tabular}
		\end{small}
	\end{center}
	\vskip -0.1in
\end{table}

\subsection{Datasets}

We choose seven classification benchmark datasets with a different scale. Table \ref{sample-table} presents the basic statistics of these datasets. The datasets vary in size: from 1484 up to 78823 instances, from 8 up to 784 features, and from 2 up to 26 classes. From the literature, these datasets come pre-divided into training and testing sets, therefore in our experiments, we use them in their original format.  PROTEIN, SENSIT, and SATIMAGE datasets are obtained from LIBSVM data sets \citep{chang11libsvm}, except for MNIST \citep{lecun98gradient} dataset, others come from the UCI Machine Learning Repository \citep{uci}. Based on the attribute characteristics of the dataset, we classify the datasets into three categories: categorical, numerical, and mixed modeling tasks. As shown above, we are expecting a better performance of the deep forest model on categorical and mixed modeling tasks.

\subsection{Configuration and Tasks}

In mdDF, we take two random forests and two completely-random forests in each layer, and each forest contains 100 trees, whose maximum depth of tree in random forest growing with the layer $t$, i.e., $d_{max}^{(t)} \in \{2t+2,4t+4,8t+8,16t+16\}$. To reduce the risk of overfitting, feature representation vector learned by each forest is generated by $k$-fold cross validation. In detail, each instance will be used as training data for $k-1$ times, produce the final class vector as \textit{augmented feature}s for the resulting in $k-1$ class vectors, which are then averaged to next layer. (See algorithm~\ref{rfb})

\begin{table}[t]
	\caption{Comparison results between mdDF and the other tree-based algorithm on test accuracy with different datasets. The best accuracy on each data set is highlighted in bold type. $\bullet$ indicates the second accuracy on each data set. The average rank is listed in the first row from the bottom.}
	\label{test1}
	\vskip -0.3in
	\begin{center}
		\begin{small}
			\begin{tabular}{p{2.0cm}p{2.0cm}<{\raggedleft}p{1.6cm}<{\centering}p{1.6cm}<{\raggedleft}p{1.6cm}<{\raggedleft}p{1.6cm}<{\raggedleft}}
				\toprule
				\bf Dataset 		&\bf MLP 		&\bf R.F. 		&\bf XGBoost	&\bf gcForest  		&\bf mdDF 			\\
				\midrule                                                        	                       
				ADULT    		&80.597			&85.818		  	&85.904			&$\bullet$86.276				&\textbf{86.560}	\\
				YEAST    		&59.641 		&61.886		  	&59.161			&$\bullet$63.004				&\textbf{63.340}	\\
				LETTER    		&96.025			&96.575		  	&95.850			&$\bullet$97.375				&\textbf{97.500}	\\
				PROTEIN	        &68.660			&68.071		  	&$\bullet$71.741			&71.590				&\textbf{71.757}	\\
				HAR         	&$\bullet$94.231 		&92.569		    &93.112			&94.224				&\textbf{94.600}	\\
				SENSIT	        &78.957			&80.133		  	&81.874			&$\bullet$82.334				&\textbf{82.534}	\\
				SATIMAGE        &91.125 		&91.200		  	&90.450			&$\bullet$91.700				&\textbf{91.750}	\\
				MNIST        	&$\bullet$98.621 		&96.831	    	&97.730			&98.252				&\textbf{98.734}	\\
				\midrule
				Avg. Rank		&3.650          &4.000          &3.750          &2.375           &1.000\\
				\bottomrule
			\end{tabular}
		\end{small}
	\end{center}
	\vskip -0.1in
\end{table}

We compare mdDF with other four common used algorithms on different datasets: multilayer perceptron (MLP), Random Forest (R.F.) \citep{breiman01random}, XGBoost \citep{chen16xgboost} and gcForest \citep{zhou17deep}. Here, we set the same number of forests as mdDF in each layer of gcForest. For random forest, we set $400\times k$ trees; and for XGBoost, we also take $400\times k$ trees. Especially, we compare them in the small sample learning task to show their resistance ability to overfitting problem.

For the multilayer perceptron (MLP) configurations, we use ReLU for activation function, cross-entropy for loss function, adadelta for optimization, no dropout for hidden layers according to the scale of training data. The network structure hyper-parameters, however, could not be fixed across tasks. Therefore, for MLP, we examine a variety of architectures on validation set, and pick the one with the best performance, then re-train the whole network on training set and report the test accuracy. The examined architectures are listed as follows: (1) input-1024-512-output;  (2) input-16-8-8-output; (3) input-70-50-output; (4) input-50-30-output; (5) input-30-20-output. 

Besides, we compare our mdDF with three other mdDF structures on different datasets: (1) the mdDF using same forests (use 4 random forests), i.e., $\text{mdDF}_{\text{SF}}$; (2) the mdDF using stacking (only transmit the prediction vectors to next layer), i.e., $\text{mdDF}_{\text{ST}}$; (3) the mdDF without ``preconc'' (only transmit the input feature vector to next layer), i.e., $\text{mdDF}_{\text{NP}}$. We hope the contrast result could demonstrate that the internal structure (different type of forests and ``preconc'' operation) of the deep forest is reasonable.

Since our method introduces the margin distribution optimization to the deep forest framework, we design an experiment to visualize the variation of boosted features over different layers of the model. In this way, we can know how our reweighting operation will benefit the in-model feature transformation in the deep forest model.

\begin{table}[t]
	\caption{Comparison results between mdDF and the other mdDF structures on test accuracy with different datasets.}
	\label{test3}
	\begin{center}
		\begin{small}
			\begin{tabular}{p{2cm}p{2cm}<{\centering}p{1.6cm}<{\centering}p{1.6cm}<{\centering}p{1.6cm}<{\centering}}
				\toprule
				\bf Dataset 	&\bf $\text{mdDF}_{\text{SF}}$ 			&\bf $\text{mdDF}_{\text{ST}}$ 		&\bf $\text{mdDF}_{\text{NP}}$  		&\bf mdDF 			\\
				\midrule                                                        	                    
				ADULT    		&86.200				&85.710		  	&85.650				&\textbf{86.560}	\\
				YEAST    		&63.000			    &62.780		  	&62.556				&\textbf{63.340}	\\
				LETTER    		&96.475				&97.300		  	&96.975				&\textbf{97.500}	\\
				PROTEIN	        &71.681				&70.291		  	&68.509				&\textbf{71.757}	\\
				HAR         	&93.926 			&94.290		    &94.060			    &\textbf{94.600}	\\
				SENSIT			&82.014 			&80.412		    &80.320			    &\textbf{82.534}	\\
				SATIMAGE        &91.600			    &91.300	  		&90.800				&\textbf{91.750}	\\
				MNIST        	&98.254				&98.101		    &98.240				&\textbf{98.734}	\\
				\bottomrule
			\end{tabular}
		\end{small}
	\end{center}
\end{table}

\subsection{Results}

\subsubsection{Test Accuracy on Benchmark datasets}

Table~\ref{test1} shows that mdDF achieves a better prediction accuracy than the other methods on several datasets. Comparing with the MLP method, deep forest models almost outperform MLP on these data sets and obtain the top 2 test accuracy on categorical or mixed modeling tasks. Thus, we conjecture that deep forest model is more suitable for categorical and mixed modeling tasks. Obviously, the deep forest model gcForest and mdDF perform better than the shallow ones, and mdDF with reweighting and boosted representations outperforms gcForest across these datasets. The empirical results show that the deep model provides an improvement in performance with in-model transformation compared to the shallow models that only have invariant features. Especially under the guidance of the margin distribution reweighting, the ensemble of different depth representations brings better generalization ability for mdDF algorithm.

Table~\ref{test3} shows that mdDF achieves a better prediction accuracy than the other mdDF structures on several datasets. The empirical results show that the internal structure of mdDF (different forests and ``preconc'' operation) is the key to achieving a better generalization performance. As we analyze in Section~\ref{theory}, the ensemble of learned features over different layer can alleviate the overfitting problem theoretically because coefficients $\alpha_{t}$s control the complexity. Different type of forests can enhance the diversity of deep forest model. According to the Eq.~(\ref{diversity}), higher diversity can help achieve a better performance.

\begin{figure}[t]
	\centering
	\centerline{\includegraphics[width=0.6\textwidth]{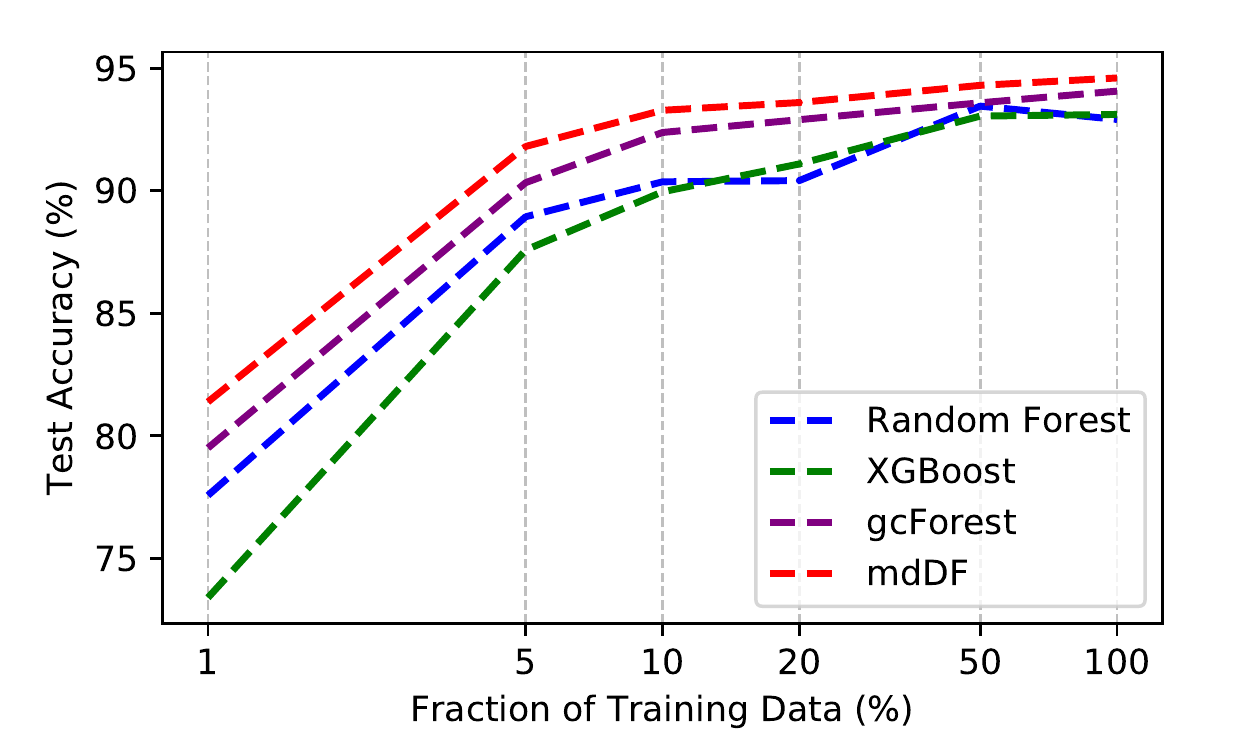}}
	\caption{ Performance of mdDF model on generalization tasks with UCI HAR dataset.  }
	\label{fig:sample1}
	\vskip -0.0in
\end{figure}

\subsubsection{Limited Training Samples}

Recently \citet{han18comparison} found the tree-based ensemble model to be more robust and stable than SVM, and neural networks, especially with small training sets. We desire that mdDF can enhance this property for tree-based model when the training data is insufficient. According to the generalization bound in Theorem~\ref{generalization}, our algorithm can restrict the complexity of the hypothesis space suitably. To evaluate the performance of mdDF model under limited training set, we randomly chose some fraction of the training set, in particular, from 100\% of the training samples to 1\% on the UCI HAR dataset, and train the models accordingly.

In Figure~\ref{fig:sample1}, we show the test accuracies of Random Forest, XGBoost, gcForest, mdDF models trained on different fractions of the UCI HAR dataset. As shown in Figure~\ref{fig:sample1}, the test accuracies of all these four models increase as the fraction of training samples increases. Obviously, the mdDF model outperforms all the other models constantly across different fractions. On the UCI HAR dataset, the mdDF model outperforms Random Forest model by around 3.87\%, XGBoost model by around 5.23\% and gcForest model by around 2.48\% on the smallest training set which contains only 1\% of the whole training samples. Thus we know that the cascade structure can efficiently improve the generalization performance on limited training set with a deep model. Especially, the optimal margin distribution principle give the cascade processing a sound explanation.

\begin{figure*}[t]
	\centering
	\begin{minipage}[htb]{\textwidth}
		\centering
		\includegraphics[width=0.95\textwidth]{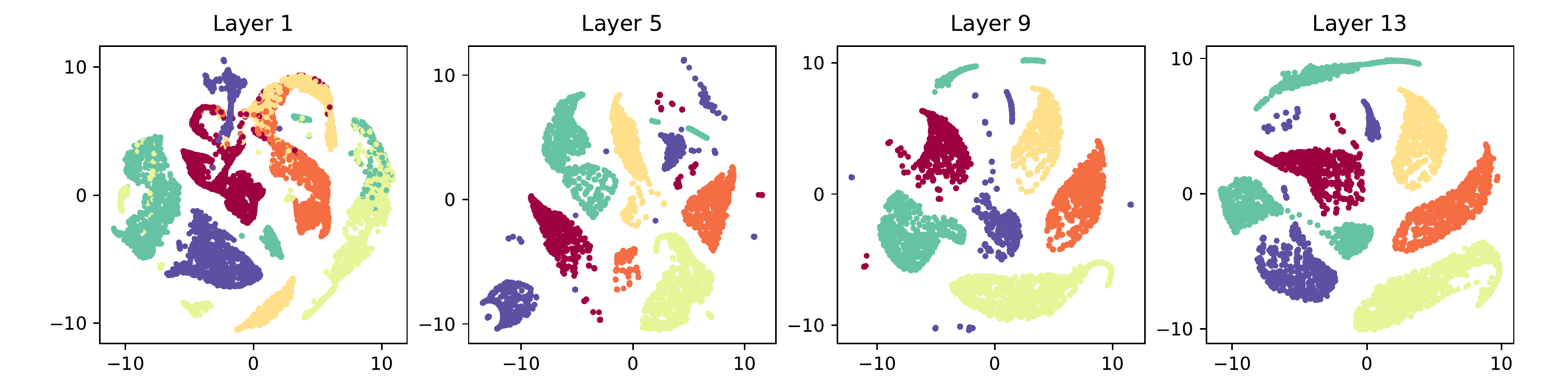}
		\centerline{(a)}
	\end{minipage}
	\begin{minipage}[htb]{\textwidth}
		\centering
		\includegraphics[width=0.95\textwidth]{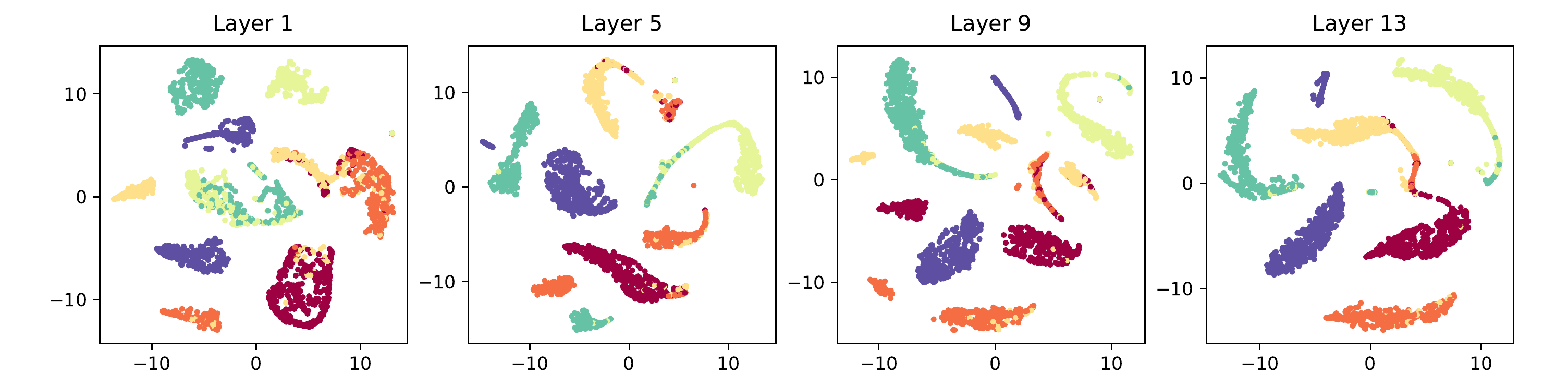}
		\centerline{(b)}
	\end{minipage}
	\caption{ Multi-layer feature visualization of mdDF on UCI HAR training (a) and test (b) data set. We do the variance decomposition in this 2D space, and the ratio of the intra-class variance to the inter-class variance $S_A/S_A$ can be obtained as follows: (a) $[3.88, 1.97, 0.72, 0.65]$ and (b) $[1.69, 0.88, 0.75, 0.51]$, i.e., the intra-class compactness and inter-class separability is getting better as the layer becomes deeper. Extensive margin distribution results are shown as a curve in Figure~\ref{fig:curve} correspondingly.}
	\label{fig:tsne}
	\vskip -0.0in
\end{figure*}

\subsubsection{Accuracy Curves over Layers}

We use our mdDF algorithm to train and test on the UCI HAR dataset plot Figure~\ref{fig:curve} based on a similar figure in \citet{schapire98boosting}. It can be observed that our method achieves $100\%$ training accuracy in less than $3$ layers, but after that, the generalization accuracy keeps increasing. As we show in Theorem.~\ref{generalization}, the margin distribution is crucial to explain why the algorithm seems resistant to overfitting problem. Especially, we can evaluate the margin distribution by calculating the ratio between the margin variance and the square of margin mean. In Figure~\ref{fig:margin_har}, we show that the margin distribution varies with the layers, that is, the margin mean becomes larger and the margin variance become smaller, and the ratio between them is shown in Figure~\ref{fig:curve}.

\subsubsection{Feature Visualization}

\begin{figure}[h]
	\vskip -0.0in
	\centering
	\begin{minipage}[h]{0.47\textwidth}
		\centering
		\includegraphics[width=\columnwidth]{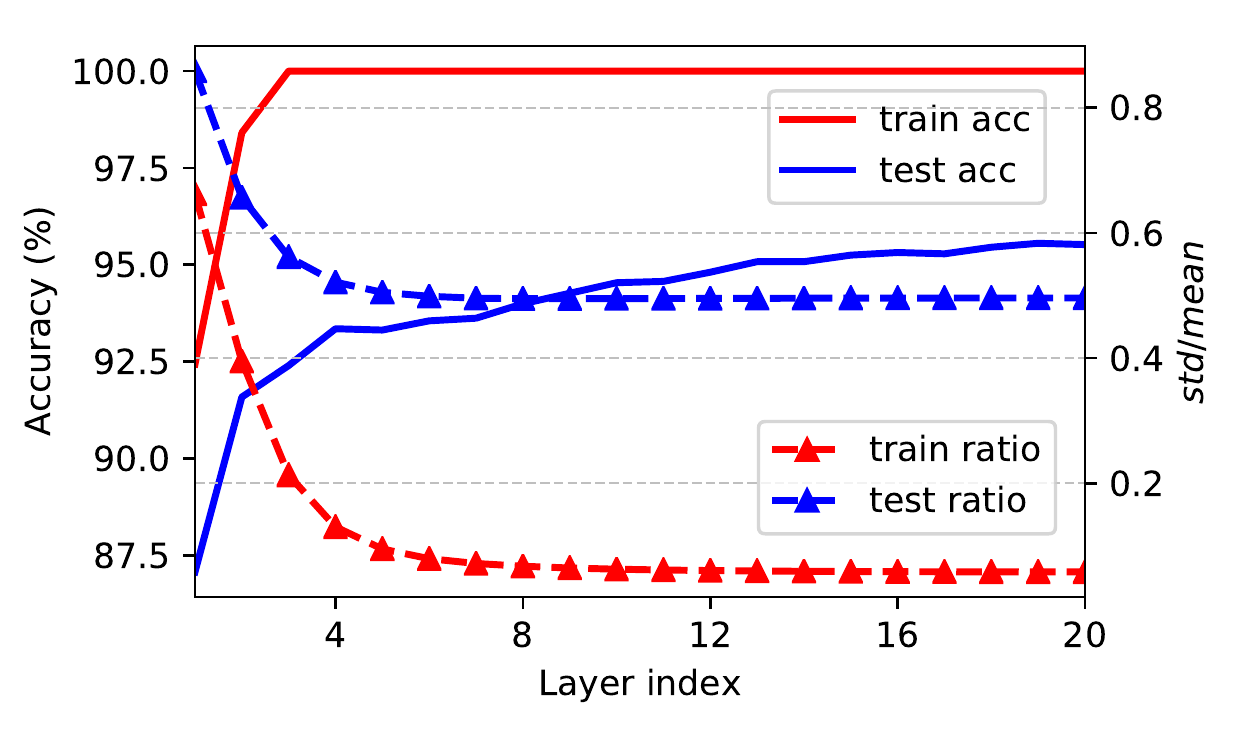}
		\caption{Training and test accuracy of mdDF model over layers on the UCI HAR data set (solid). Margin rate over layers on the UCI HAR data set (dot-dashed).}
		\label{fig:curve}
	\end{minipage}
	\begin{minipage}[h]{0.04\textwidth}
		\includegraphics[width=\columnwidth]{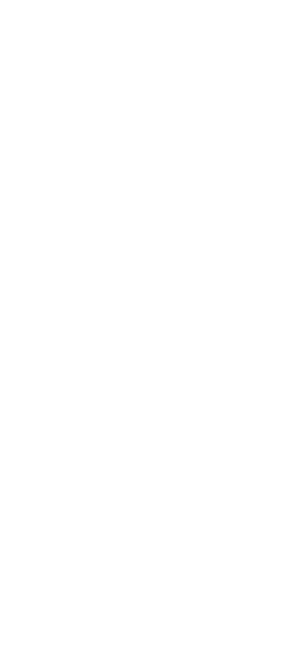}
	\end{minipage}
	\begin{minipage}[h]{0.47\textwidth}
		\centering
		\includegraphics[width=\columnwidth]{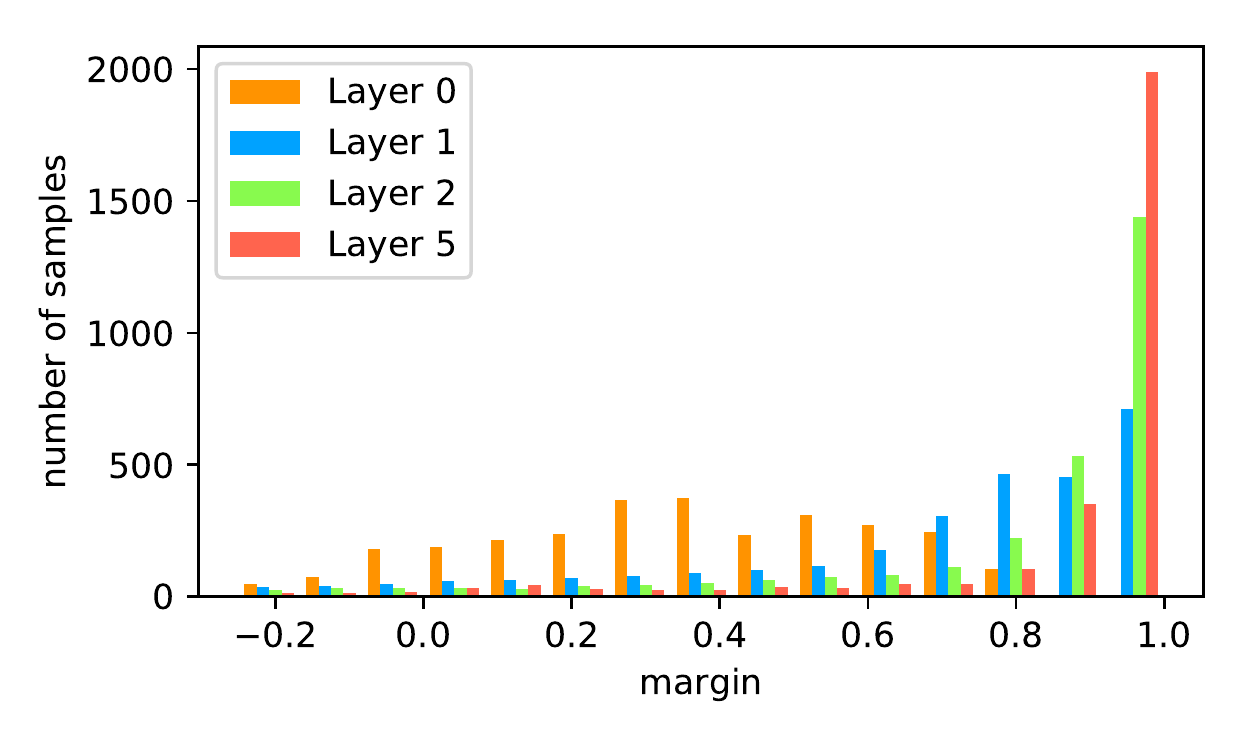}
		\caption{ Margin distribution of mdDF model on the UCI HAR data set.}
		\label{fig:margin_har}
	\end{minipage}
	\vskip -0.2in
\end{figure}


Since the performance of the mdDF model is excellent, we hope to see that the distributions of data in the learned feature space (over different layers) are consistent with the generalization ability. In this experiment, we use the t-SNE method to visualize the data distribution over different layers for training samples and test samples.
Figure~\ref{fig:tsne} plots the 2-dimension embedding image on UCI HAR dataset.
The t-SNE \citep{maaten08visualizing} is a tool to visualize high-dimensional data. It converts similarities between data points to joint probabilities and tries to minimize the Kullback-Leibler divergence between the joint probabilities of the low-dimensional embedding and the high-dimensional data.

Consistently, we can find that the visualization of our model is getting better as the layer becomes deeper, the distribution of the samples which has the same label is more compact. To quantify the degree of compactness of the distribution, we perform a variance decomposition on the data in the embedding space. Compared with the ratio of the standard deviation of margin to the mean of margin in Figure~\ref{fig:curve}, we can know that mdDF model attains a better distribution with a small ratio $\lambda$ while the layer becomes deeper.

\section{Conclusion} \label{conc}

Recent studies propose a few tree-based deep models to learn the representations from a broad range of tasks and achieve good performance. We extend the margin distribution theory to explain how the deep forest models learn the different representations over layers and further guide the ensemble method to obtain a better performance. We also propose the mdDF approach, which aims to get a better margin distribution, i.e., a small ratio $\lambda$. As for experiments, the results validate the superiority of our method on different datasets and show its powerful ability to learn good representations.


\bibliographystyle{plainnat}
\bibliography{example_paper}

%
%
%
%

\end{document}